%% file: main.tex
\documentclass[letterpaper, 10 pt, conference]{ieeeconf}  

\IEEEoverridecommandlockouts                     

\title{\ours : Composable Language-Annotated Whole-body Motion Generation}

\author{
    {Jianuo Cao}$^{\ast,1,2}$, {Yuxin Chen}$^{\ast,2}$, {Masayoshi Tomizuka}$^{2}$\\
    $^{\ast}$Denotes equal contribution \\
    $^1$\textit{Nanjing University} \quad $^2$\textit{University of California, Berkeley}\\
}

\input{notion_marcos}

\begin{document}
\maketitle
\thispagestyle{empty}
\pagestyle{empty}
\begin{strip}
    \centering
    \vspace{-25pt} 
    \includegraphics[width=\textwidth, height=10cm]{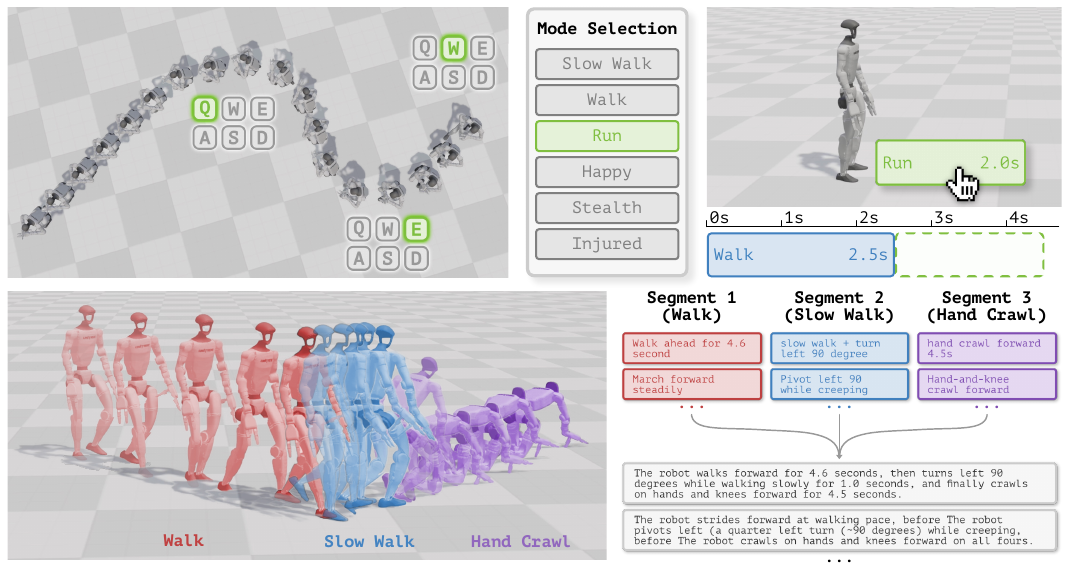} 
    \captionof{figure}{\textbf{Overview of \ours.} Users can compose whole-body motion sequences for the Unitree G1 humanoid through either keyboard mode \textbf{(top-left)} or editor mode \textbf{(top-right)}, producing diverse trajectories with seamless transitions between motion modes \textbf{(bottom-left)}. Each trajectory is automatically paired with multi-style natural-language descriptions \textbf{(bottom-right)}.}
    \label{fig:teaser}
    \vspace{0pt} 
\end{strip}

\begin{abstract}

Training language-conditioned whole-body controllers for humanoid robots demands large-scale motion–language datasets. Existing approaches based on motion capture are costly and limited in diversity, while text-to-motion generative models produce purely kinematic outputs that are not guaranteed to be physically feasible. We present \ours, a pipeline for scalable generation of language-annotated whole-body motion data for the Unitree G1 humanoid robot. \ours composes motion primitives from a kinematic planner, parameterized by movement, heading, speed, pelvis height, and duration, and provides two browser-based interfaces---a real-time keyboard mode and a timeline-based sequence editor---for exploratory and batch data collection. A low-level controller tracks these references in MuJoCo simulation, yielding physically grounded trajectories. In parallel, a template-based engine generates diverse natural-language annotations at both segment and trajectory levels.
To support scalable generation of motion-language paired data for humanoid robot learning, we make our system publicly available at: \url{https://github.com/JianuoCao/CLAW}
\end{abstract}

\input{sections/intro}
\input{sections/method}
\input{sections/results}
\input{sections/related_work}
\input{sections/conclusion}

\bibliographystyle{IEEEtran}
\bibliography{reference}

\end{document}

%% file: notion_marcos.tex
\usepackage{amsmath,amssymb,stmaryrd,mathtools}
\usepackage{xcolor}
\usepackage{xspace}
\usepackage{bbm}
\usepackage{cuted}
\usepackage{caption}
\usepackage{subcaption}
\usepackage{graphicx}
\usepackage{tcolorbox}
\usepackage{multicol}
\usepackage[bookmarks=true]{hyperref}
\usepackage{colortbl}
\usepackage{multirow}
\usepackage{booktabs} 
\usepackage{wrapfig}
\usepackage{lipsum}
\usepackage{float}
\usepackage{cleveref}
\definecolor{dark_red}{RGB}{122, 0, 0}
\definecolor{coral}{RGB}{255, 119, 94}
\definecolor{pink_orange}{RGB}{255, 72, 126}
\definecolor{vibrant_pink}{RGB}{255, 0, 104}
\definecolor{pink_pink}{RGB}{255, 37, 153}
\definecolor{wine}{RGB}{204, 0, 102}

\definecolor{light_orange}{RGB}{255, 198, 107}
\definecolor{orange(sae/ece)}{rgb}{1.0, 0.49, 0.0}
\definecolor{dark_orange}{RGB}{216,92,0}

\definecolor{org-purp-0}{RGB}{165, 76, 0}
\definecolor{org-purp-1}{RGB}{250, 130, 28}
\definecolor{org-purp-2}{RGB}{226, 89, 68}
\definecolor{org-purp-3}{RGB}{206, 92, 124}
\definecolor{org-purp-4}{RGB}{116, 80, 146}
\definecolor{org-purp-5}{RGB}{110, 78, 157}

\definecolor{teal(sae/ece)}{rgb}{0, 0.47, 0.52}
\definecolor{aqua}{RGB}{52,172,139}
\definecolor{dark_aqua}{RGB}{35,115,93}
\definecolor{dark_green}{RGB}{0, 92, 34}

\definecolor{grape}{RGB}{112,48,160}
\definecolor{purple}{rgb}{0.74, 0.65, 1.0}
\definecolor{dark_purple}{rgb}{0.58, 0.0, 0.82}
\definecolor{periwinkle}{RGB}{191, 140, 230}

\definecolor{light_gray}{rgb}{0.9, 0.9, 0.9}
\definecolor{medium_gray}{rgb}{0.6, 0.6, 0.6} 
\definecolor{dark_gray}{rgb}{0.2, 0.2, 0.2} 

\definecolor{sky_blue}{RGB}{37, 166, 213}
\definecolor{light_blue}{rgb}{0.33, 0.80, 1}
\definecolor{dark_blue}{rgb}{0.098, 0.239, 0.52}
\definecolor{ocean}{RGB}{13, 121, 202}
\definecolor{light_ocean}{RGB}{18, 178, 235}
\definecolor{dark_ocean}{RGB}{10, 89, 148}
\definecolor{vibrant_blue}{RGB}{14, 120, 255}

\definecolor{dark_brown}{rgb}{0.3255, 0.004, 0.001}

\newcommand{\para}[1]{\medskip\noindent\textbf{#1. }}

\newcounter{qnum}
\newcounter{tnum}
\newcommand{\ours}{\textcolor{dark_gray}{\texttt{CLAW}}\xspace}

\usepackage{amsmath}


%% file: sections/intro.tex
\section{Introduction}
\begin{figure*}
    \centering
    \includegraphics[width=0.9\linewidth]{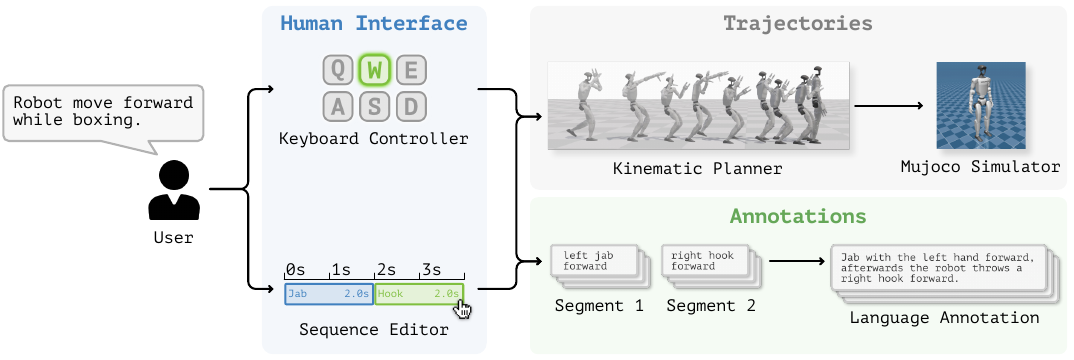}
    \caption{\textbf{Overview of the \ours data generation pipeline.} A human operator specifies motion intent via a keyboard controller or sequence editor. Structured commands are streamed to a kinematic planner whose reference motions are tracked by a whole-body controller in MuJoCo simulation. A template-based annotation engine simultaneously produces diverse natural-language descriptions from the same motion parameters, yielding time-aligned trajectory and language data.}
    \label{fig:data_gen}
\end{figure*}

High-quality reference motions are essential for training humanoid robots to perform whole-body loco-manipulation tasks.
The prevailing paradigm acquires such data through human motion capture (MoCap) followed by kinematic retargeting to the target morphology~\cite{luo2023perpetual,peng2018deepmimic}.
While effective, this pipeline faces three fundamental bottlenecks:
(i)~MoCap sessions are expensive and logistically constrained, bounding both the volume and behavioral diversity of the resulting corpus;
(ii)~retargeting across disparate embodiments introduces kinematic artifacts---foot sliding, self-penetration, and joint-limit violations---that degrade downstream policy performance~\cite{araujo2025retargeting};
and (iii)~every recorded clip must be manually paired with a natural-language description when the target task requires language conditioning, creating an annotation cost that grows linearly with dataset size.

Recent advances in generative motion models have begun to address the data scarcity problem.
Large-scale MoCap datasets such as AMASS~\cite{mahmood2019amass} have greatly expanded the
coverage of human motion, providing the foundation upon which text-to-motion diffusion models
like Kimodo~\cite{rempe2026kimodo} are trained to synthesize diverse whole-body movements from language prompts.

However, these approaches operate in a purely kinematic regime: the generated motions are not guaranteed to be physically feasible, and an additional physics-based tracking step~\cite{luo2023perpetual} is typically required before deployment on a real robot.
Moreover, the coupling between text input and motion output makes it difficult to independently control meta motion attributes such as speed, heading, pelvis height and gait style---properties that are critical for systematic data augmentation.

In this work, we take a complementary approach: rather than generating motions \emph{from} language, we produce language-annotated motion data \emph{at scale} by composing parametric motion primitives in physics simulation.
Our key observation is that a kinematic planner paired with a low-level whole-body controller already produces diverse, physically grounded motions across a broad behavioral repertoire, yet exposes only low-level command interfaces that are unsuitable for systematic dataset construction.
We bridge this gap by treating the planner's motion modes as composable primitives, each governed by parameters like movement, heading, speed, pelvis height and duration.
This compositional abstraction allows arbitrarily long and diverse trajectory sequences to be assembled through simple parameterization, entirely without motion-capture infrastructure.

We present \ours (\cref{fig:teaser}), an interactive web-based pipeline for scalable generation of language-annotated whole-body motion data for the Unitree G1 humanoid robot.
As illustrated in~\cref{fig:data_gen}, \ours exposes two complementary interfaces---a real-time keyboard mode for free-form exploration and a timeline-based sequence editor for reproducible multi-segment recipes---both streaming commands at 20\,Hz to a kinematic planner via a WebSocket--ZMQ bridge.
The planner's kinematic references are tracked by a low-level whole-body controller in MuJoCo, yielding physically grounded trajectories recorded at 50\,Hz.
In parallel, a deterministic template-based annotation engine converts the same motion parameters into diverse natural-language descriptions(see~\cref{sec:annotation}).

We instantiate the current pipeline with the kinematic planner and whole-body controller from SONIC~\cite{luo2025sonicsupersizingmotiontracking}; however, \ours communicates with the controller through a generic command interface and is agnostic to the specific backend.

The key contributions of this report are:
\begin{itemize}
    \item A composable data-generation pipeline that converts parameterized motion primitives into physically simulated, language-annotated whole-body trajectories for the Unitree G1, with a planner-agnostic architecture that decouples data collection from the choice of motion-generation backend.
    \item Two interactive browser-based interfaces---keyboard and sequence editor---that enable both exploratory and batch data collection through a unified backend.
    \item A template-based language annotation engine that deterministically produces multiple stylistic descriptions per trajectory, providing diverse yet reproducible supervision for language-conditioned control.
    \item An open-source, four-process system architecture (simulation, controller, WebSocket bridge, browser frontend) that is modular, extensible, and capable of generating training data at arbitrary scale.
\end{itemize}

%% file: sections/method.tex
\section{\ours Pipeline}
As illustrated in~\cref{fig:data_gen}, the \ours pipeline produces large-scale, language-annotated whole-body motion data for the Unitree G1 humanoid robot.
A human operator specifies motion intent---comprising motion mode, movement direction, heading, speed, pelvis height and duration ---through one of two interactive interfaces: a real-time keyboard mode or a sequence editor mode.
These structured commands are sent to a kinematic planner, which synthesizes short-horizon reference motions.
A low-level whole-body controller then tracks these references in MuJoCo simulation, producing physically consistent trajectories recorded at 50\,Hz.
In parallel, the same motion-intent parameters are fed to a template-based language annotation engine that generates diverse natural-language descriptions, both per-segment and for the full trajectory.
The final output for each session is a structured data package containing time-aligned kinematic and dynamic trajectory and multiple stylistic language annotations.
\begin{figure*}[t!]
    \centering
    \begin{subfigure}{0.495\linewidth}
        \centering
        \includegraphics[width=\linewidth]{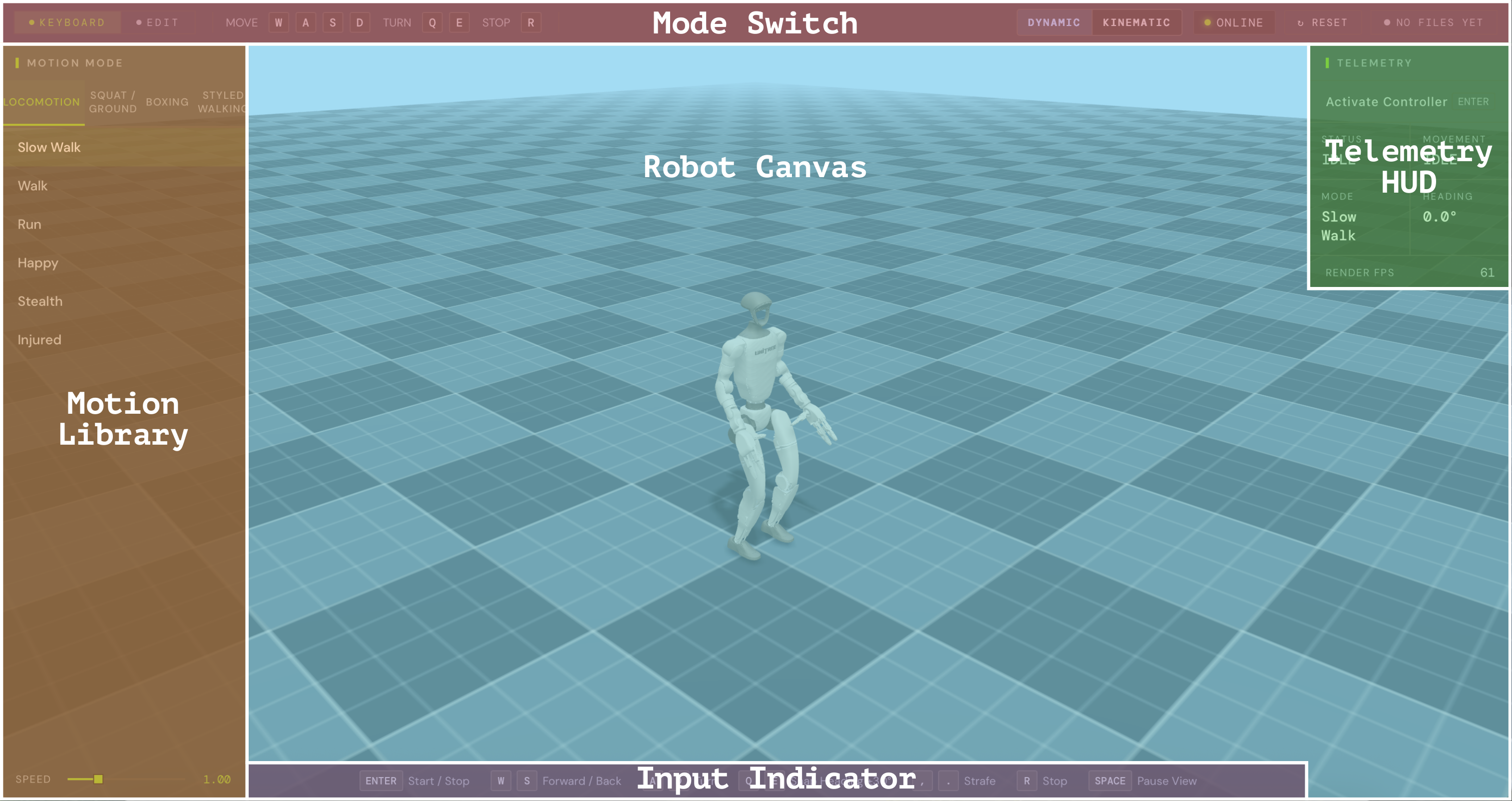}
        \caption{Keyboard Mode}
        \label{fig:ui_left}
    \end{subfigure}
    \hfill 
    \begin{subfigure}{0.495\linewidth}
        \centering
        \includegraphics[width=\linewidth]{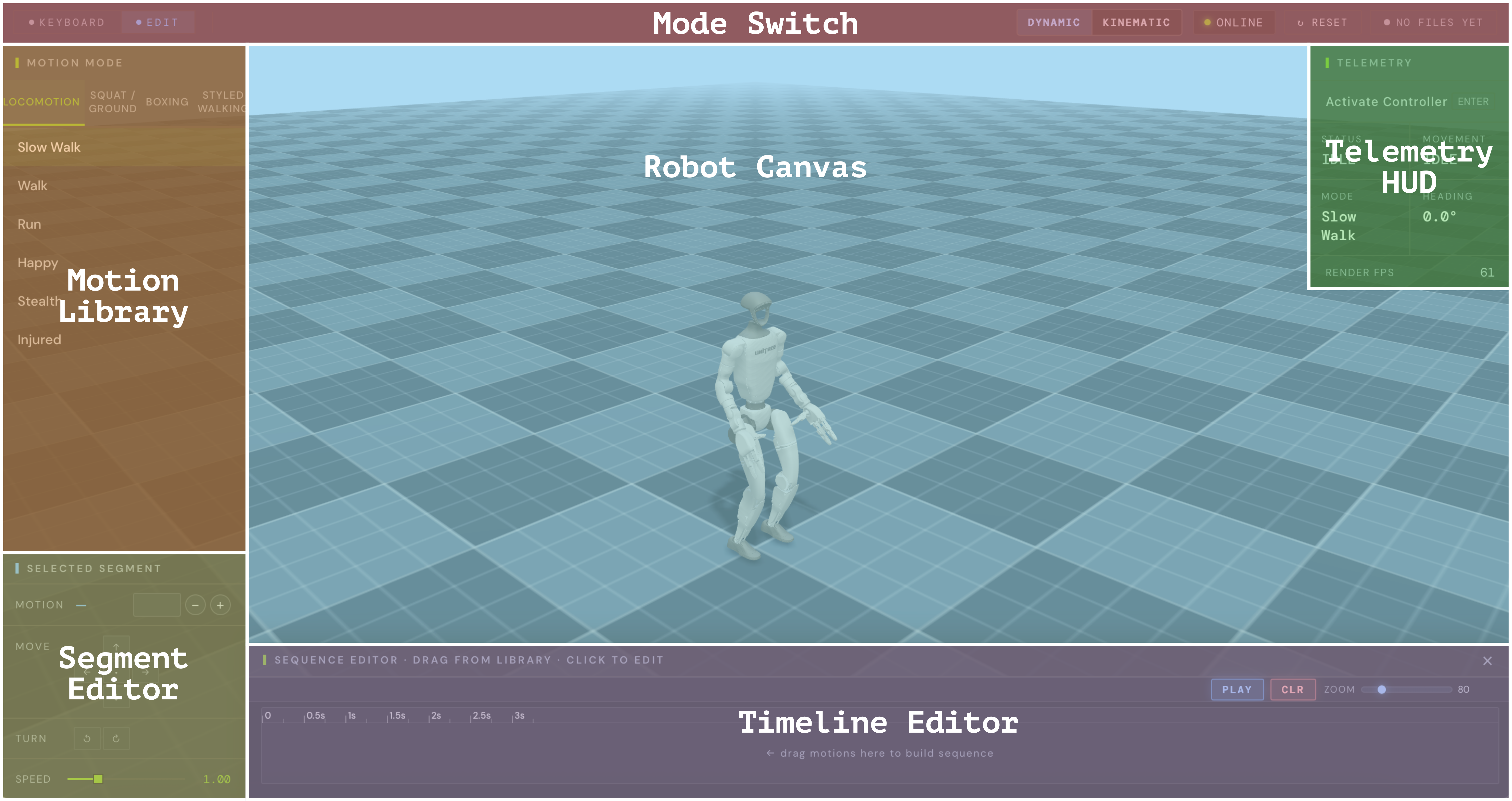}
        \caption{Editor Mode}
        \label{fig:ui_right}
    \end{subfigure}
    
    \caption{\textbf{Two complementary control interfaces of \ours.} (a)~In keyboard mode, the operator controls the robot interactively using key bindings for motion mode, movement, heading, speed, pelvis height and duration. (b)~In editor mode, motion clips are arranged on a visual timeline with per-segment configuration, enabling reproducible large-batch data generation.}
    \label{fig:user_interface}
\end{figure*}

\subsection{User Interface}

The first stage of the pipeline is the user interface, which translates operator intent into the structured commands consumed by the planner.
\ours exposes two complementary control modes, both streaming commands at 20\,Hz to the same backend (\cref{fig:user_interface}).

\para{Keyboard Mode}
In keyboard mode (\cref{fig:ui_left}), the operator controls the robot interactively in real time.
The \texttt{W/A/S/D} keys set the movement direction (forward, turn-left, backward, turn-right), \texttt{Q/E} snap the heading by $\pm30^\circ$, and the comma and period keys trigger lateral strafing; pressing \texttt{R} immediately halts all movement.

A motion-mode selector in the left panel organizes the 25 available behaviors into four categories---Locomotion, Ground, Boxing, and Styled Walking---while a speed slider and a pelvis-height slider allow continuous adjustment within the range supported by the active mode.

A telemetry panel on the right displays the current controller status, movement type, active mode, heading angle, and rendering frame rate.
Because every keypress is translated into a planner command within a single control cycle, keyboard mode is well suited for free-form exploration and rapid prototyping of novel motion sequences.

\para{Editor Mode}
In editor mode (\cref{fig:ui_right}), the operator composes multi-segment trajectories on a visual timeline.
Motion clips are dragged from the library into the sequence editor, where each segment can be individually configured with a target duration, movement direction, optional turn angle, speed override, and---for ground-level modes---pelvis height.

Once a sequence is finalized, a single click dispatches it to the backend for execution: the bridge iterates through the segments, drives the planner, records both kinematic and dynamic trajectories at 50\,Hz.

This mode is particularly valuable for reproducible, large-batch data generation, as identical recipes can be replayed with varied parameters to systematically expand the training corpus.

\subsection{Kinematic Planner and Motion Primitives}

\ours treats the kinematic planner as a black-box motion-primitive engine.
At each control step the bridge sends a structured meta command over ZMQ containing five fields: a motion-mode index, a movement-direction vector, a facing-direction vector, a speed scalar, and a pelvis-height scalar.
The planner consumes this command together with the current robot state and produces a kinematic reference motion, which is then tracked by a low-level whole-body controller in MuJoCo simulation to yield physically grounded joint trajectories.

The planner supports 25 motion modes organized into four groups---Locomotion, Squat/Ground, Boxing, and Styled Walking---as summarized in~\cref{tab:motion_mode}.
Each mode accepts a different subset of continuous parameters.This parameterization renders each mode a self-contained motion primitive that can be composed with any other mode to form longer sequences.

Composability arises from mode switching: when the bridge sends a new mode index mid-session, the resulting motion transitions smoothly from the current state to the new behavior without any explicit blending or post-processing on the pipeline side.
This property enables heterogeneous primitives. For example, walking into a squat followed by a crawl, to be chained into continuous trajectories.

In this work we use the kinematic planner and whole-body controller from SONIC~\cite{luo2025sonicsupersizingmotiontracking}; see the original work for architectural and performance details.
\begin{table}[htbp]
    \centering
    \caption{Available motion modes exposed by the planner. Checkmarks indicate whether a mode supports adjustment of speed, heading or height.}
    \label{tab:motion_mode}
    \footnotesize
    \setlength{\tabcolsep}{5pt}
    \begin{tabular}{llccc}
        \toprule
        \textbf{Group} & \textbf{Mode Name} & \textbf{Speed} & \textbf{Heading} & \textbf{Height} \\
        \midrule
        
        \multirow{6}{*}{Locomotion}
        & Slow Walk        & \checkmark & \checkmark &  \\
        & Walk             &  & \checkmark &  \\
        & Run              & \checkmark & \checkmark &  \\
        & Happy            &  & \checkmark &  \\
        & Stealth          &  & \checkmark &  \\
        & Injured          &  & \checkmark &  \\
        
        \midrule
        
        \multirow{5}{*}{Squat/Ground}
        & Squat            &  &  & \checkmark \\
        & Kneel (Two)      &  &  & \checkmark \\
        & Kneel (One)      &  &  & \checkmark \\
        & Hand Crawl       & \checkmark & \checkmark & \checkmark \\
        & Elbow Crawl      & \checkmark & \checkmark & \checkmark \\
        
        \midrule
        
        \multirow{7}{*}{Boxing}
        & Idle Boxing      &  & \checkmark &  \\
        & Walk Boxing      & \checkmark & \checkmark &  \\
        & Left Jab         & \checkmark & \checkmark &  \\
        & Right Jab        & \checkmark & \checkmark &  \\
        & Random Punches   & \checkmark & \checkmark &  \\
        & Left Hook        & \checkmark & \checkmark &  \\
        & Right Hook       & \checkmark & \checkmark &  \\
        
        \midrule
        
        \multirow{7}{*}{Styled Walking}
        & Careful          &  & \checkmark &  \\
        & Object Carrying  &  & \checkmark &  \\
        & Crouch           &  & \checkmark &  \\
        & Happy Dance      &  & \checkmark &  \\
        & Zombie           &  & \checkmark &  \\
        & Point            &  & \checkmark &  \\
        & Scared           &  & \checkmark &  \\
        
        \bottomrule
    \end{tabular}
\end{table}

\begin{figure*}
    \centering
    \includegraphics[width=\linewidth]{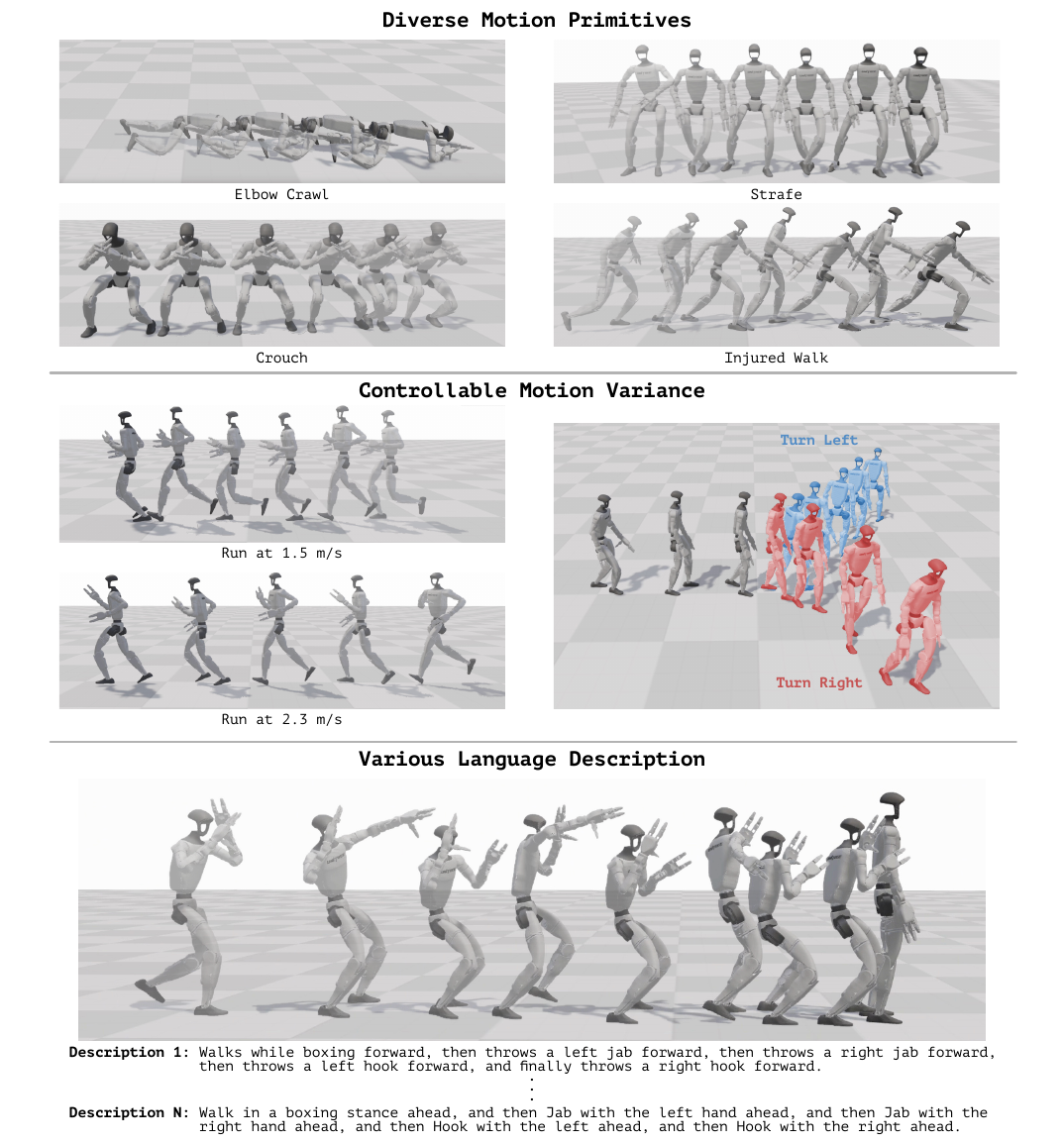}
    \caption{\textbf{Illustrative examples of \ours generation capabilities.} (a)~The pipeline produces diverse whole-body motions spanning locomotion, squatting, boxing, and styled walking. (b)~For a given motion mode, parameters such as velocity and heading can be continuously adjusted. (c)~Each generated motion sequence is automatically annotated with multiple stylistically varied natural-language descriptions.}
    \label{fig:method}
\end{figure*}
\subsection{Language Annotation}
\label{sec:annotation}
Once a trajectory has been recorded, \ours automatically produces natural-language descriptions for every segment and for the trajectory as a whole(\cref{fig:method}).
The annotation engine operates on the same motion-intent parameters that drive the planner---motion mode, movement direction, heading, speed and duration---and converts them into fluent sentences through a template-based rendering pipeline.

Eight distinct annotation styles are supported, organized along two orthogonal factors.
The first factor controls register: \emph{instruction} (imperative, e.g.\ ``Walk forward for 3.0 seconds''), \emph{natural} (adverbial, e.g.\ ``Stride ahead briskly for about 3.0 seconds''), \emph{narrative} (third-person, e.g.\ ``The robot marches forward for 3.0 seconds''), and \emph{concise} (keyword-only, e.g.\ ``walk forward 3.0s'').
The second factor controls temporal grounding, with each register rendered both with and without explicit duration, resulting in eight variants for every motion segment.

Linguistic diversity is achieved through synonym banks rather than stochastic sampling from a language model.
Each of the 25 motion modes is associated with a verb bank of synonyms (e.g.\ \texttt{run} $\rightarrow$ \{run, sprint, dash, jog quickly, move at full speed\}), and analogous banks exist for movement directions and turn verbs.
Speed values are mapped to tempo adverbs via linear interpolation over per-mode speed ranges (e.g.\ for \texttt{run}, adverbs range from ``at a jog'' to ``at full speed'' across 1.5--3.0\,m/s), and turn magnitudes are discretized into human-friendly buckets (slight $<15^\circ$, partial $<60^\circ$, quarter $<120^\circ$, half $<240^\circ$).
Consecutive segments are joined with varied connectives drawn from a bank of options such as
``then'' ``followed by'' ``after which'' ``next''.


Because the generation process is purely deterministic given a seed, annotations are exactly reproducible while still exhibiting substantial lexical variety across sessions, striking a practical balance between diversity and consistency for large-scale supervised training of language-conditioned motion models.
Importantly, this design does not preclude the use of large language models. The proposed pipeline is modular and can be naturally extended to incorporate LLM-based annotation when greater linguistic diversity or richer paraphrastic variation is desired.

\begin{figure*}
    \centering
    \includegraphics[width=\linewidth]{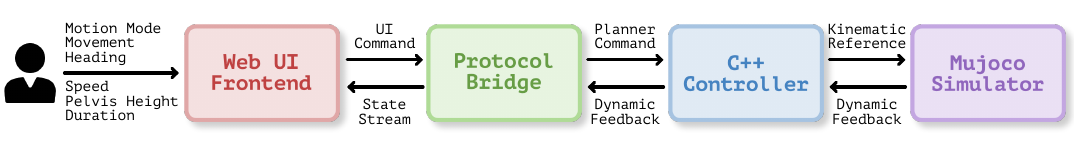}
    \caption{\textbf{System architecture of \ours.} The pipeline comprises four decoupled processes: a MuJoCo simulation, a C++ controller hosting the kinematic planner and tracking policy, a WebSocket--ZMQ bridge for protocol translation and recipe orchestration, and a browser-based frontend for visualization and operator interaction.}
    \label{fig:sys_diag}
\end{figure*}
\subsection{System Design}

The \ours application is realized as a four-process distributed system (\cref{fig:sys_diag}), in which each process runs independently and communicates through well-defined message channels.

The web UI frontend serves as the user interaction layer, allowing operators to specify high-level motion parameters, including motion mode, movement direction, heading, speed, pelvis height, and duration. These inputs are issued as UI commands at 20Hz and streamed to the backend, while system states are visualized in real time via a return state stream.

The protocol bridge, implemented as a WebSocket–ZMQ interface, acts as the central coordination layer. It translates communication protocols, maintains the planner state, aggregates telemetry into a unified state representation, and orchestrates command sequencing and trajectory management. The bridge forwards planner commands to the controller via ZMQ at 20Hz.

The C++ controller performs kinematic planning and whole-body tracking. It converts high-level planner commands into kinematic references and joint-level targets, executes control policies, and returns telemetry (including joint states and base motion) to the bridge at approximately 50Hz.

The MuJoCo simulator provides the physics execution environment, advancing system dynamics at the control rate using the Unitree G1 model and returning state feedback in response to control inputs.

%% file: sections/results.tex
\begin{figure}[htbp]
    \centering
    \begin{subfigure}{\linewidth}
        \centering
        \includegraphics[width=\linewidth]{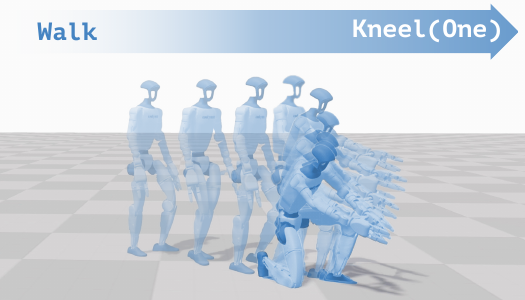}
        \caption{Successful transition}
        \label{fig:stitching_success}
    \end{subfigure}
    \hfill 
    \begin{subfigure}{\linewidth}
        \centering
        \includegraphics[width=\linewidth]{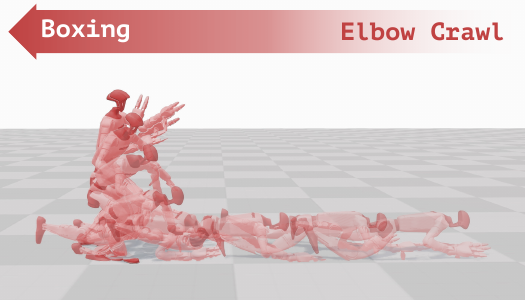}
        \caption{Failure transition}
        \label{fig:stitching_failure}
    \end{subfigure}
    
    \caption{\textbf{Motion stitching.} \ours enables motion stitching across semantically distinct motion modes. (a)~The generative planner produces smooth, natural transitions between different motion modes. (b)~Abrupt mode switching might lead to transient artifacts in the generated motion.}
    \label{fig:motion_stitching}
\end{figure}
\section{Key Results}

We demonstrate three key capabilities of \ours.
First, the composable motion primitive abstraction enables smooth stitching of semantically distinct behaviors into long-horizon trajectories, with the planner handling transitions automatically.
Second, the automated language annotation engine produces accurate, stage-wise natural-language descriptions for multi-segment sequences.
Third, the pipeline scales to arbitrary dataset sizes: the editor interface supports reproducible batch generation of motion--language pairs, and because the planner operates natively on the Unitree G1 skeleton, the resulting trajectories require no retargeting step.

\begin{figure*}
    \centering
    \includegraphics[width=0.9\linewidth]{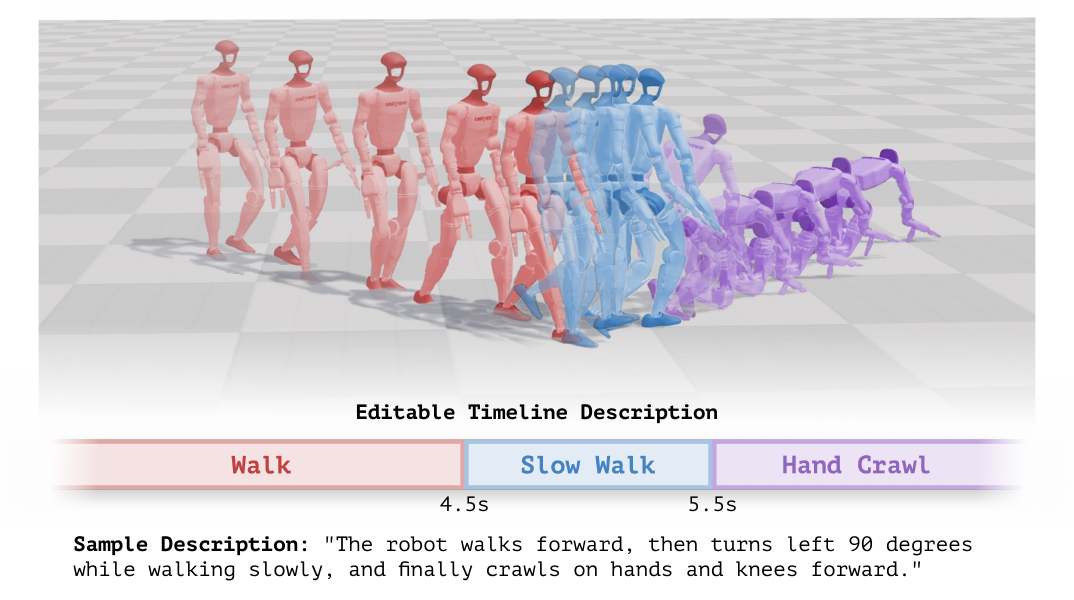}
    \caption{\textbf{Multi-stage trajectory.} Example of automated language annotation for a long-horizon trajectory with multiple stages.}
    \label{fig:auto_annotation}
\end{figure*}
\subsection{Smooth Motion Stitching}

A central feature of \ours is composing motion primitives with semantically distinct behaviors into continuous trajectories.
As described in the method section, the planner uses the current robot state as its starting point whenever the active mode changes, producing smooth transitions without any explicit blending on \ours's side.
As illustrated in~\cref{fig:motion_stitching}, the majority of motion-mode pairs achieve natural-looking transitions.
\cref{fig:stitching_success} presents a representative success case, in which the robot transitions fluidly across locomotion modes that differ substantially in speed and body configuration, with no perceptible discontinuity in either pose or velocity.
However, for a small number of motion pairs involving extreme postural differences, the planner may produce transitions that are kinematically abrupt.
\cref{fig:stitching_failure} demonstrates one such case, in which a transition from a ground-level crawling motion to an upright boxing motion results in an unnatural intermediate posture.
Identifying and filtering such cases in large-scale data collection remains an open problem.

    

\subsection{Automated Language Annotation}

Providing accurate natural-language annotations for large-scale motion datasets is labor-intensive when performed manually.
As described in the method section, \ours addresses this through a template-based annotation engine that operates on the same structured motion-intent parameters used to drive the planner, producing stage-wise text labels for motion sequences of arbitrary length without human effort.
As shown in~\cref{fig:auto_annotation}, given a long-horizon trajectory composed of multiple motion-primitive segments, the engine produces 17 full-trajectory descriptions and eight per-segment annotations that capture the semantic content of each stage---for example, a standard walking gait transitioning into a slow left turn, and finally into a ground-level crawl.
Because the annotations are generated deterministically from the same structured parameters that drive the planner, they are aligned with the recorded segments by construction.

\subsection{Infinite Data Synthesis for Real-World Execution}
A key advantage of \ours over MoCap-based pipelines is that data collection scales with compute rather than physical recording sessions.
The editor interface makes this practical: a user defines a recipe once, and the pipeline executes, records, and annotates it without further intervention.
Identical recipes can be replayed with different speed or heading parameters to systematically vary the resulting trajectories, and new recipes can be composed from the 25 available motion primitives to cover previously unseen behavior combinations.

For each recording session, \ours produces two complementary forms of trajectory data: kinematic reference trajectories (the planner's output) and dynamic trajectories (the MuJoCo-simulated response to those references).
Because the planner operates natively on the Unitree G1 skeleton, no retargeting step is needed---the recorded joint trajectories are directly in the robot's configuration space.

%% file: sections/related_work.tex


%% file: sections/conclusion.tex
\section{Conclusion}

We presented \ours, an interactive web-based pipeline for generating language-annotated whole-body motion data for the Unitree G1 humanoid robot.
By treating the motion modes of a kinematic planner as composable primitives, \ours enables the construction of diverse, long-horizon trajectory sequences through simple parameterization.
A template-based annotation engine produces eight per-segment and seventeen full-trajectory natural-language descriptions deterministically from the same structured parameters that drive the planner, eliminating the need for manual labeling.
The four-process system architecture--- browser frontend, WebSocket--ZMQ bridge, controller and simulation ---is modular and planner-agnostic, allowing the motion-generation backend to be replaced without modifying the rest of the pipeline.

Several directions remain for future work.
First, the current language annotations are generated from templates; integrating a language model could increase stylistic diversity and produce more natural descriptions, though at the cost of determinism.
Second, while the planner supports 25 motion modes, the set is fixed; extending \ours to accept user-defined primitives or learned skills would broaden the range of generated behaviors.
Finally, automatically identifying and filtering low-quality transitions between certain motion-mode pairs would improve the reliability of large-scale batch generation.
